\documentclass[11pt]{article}

\usepackage[final]{acl}

\usepackage{times}
\usepackage{latexsym}

\usepackage[T1]{fontenc}
\usepackage[utf8]{inputenc}

\usepackage{microtype}

\usepackage{inconsolata}

\usepackage{graphicx}

\usepackage{amsmath}
\usepackage[table]{xcolor}
\usepackage{colortbl}
\usepackage{amssymb}
\usepackage{graphicx}
\usepackage{booktabs} % for \toprule, \midrule, \bottomrule
\usepackage[table]{xcolor} % for \rowcolor
\usepackage{multirow}
\usepackage{float}
\usepackage{stfloats}
\usepackage{xurl}
\usepackage{tcolorbox}
\usepackage{xcolor}
\tcbuselibrary{skins,listings,breakable}
\newtcblisting{promptbox}[1]{
    enhanced,
    breakable,
    colback=white,
    colframe=gray!70,
    colbacktitle=white,
    coltitle=black,
    boxrule=0.6pt,
    arc=2pt,
    title=#1,
    fonttitle=\bfseries\small,
    halign title=center,
    left=5pt,right=5pt,top=4pt,bottom=4pt,
    listing only,
    listing options={
        basicstyle=\ttfamily\footnotesize,
        breaklines=true,
        breakindent=0pt,
        columns=fullflexible,
        keepspaces=true,
        frame=none
    }
}
\usepackage{fontawesome5}

\newcommand{\methodname}{\textsc{StaIF}} 
\newcommand{\datasetname}{\textsc{StaInstruct}}  % 

\title{\methodname: A Stage-wise Optimization for Complex Instruction Following}

\author{
  \textbf{Jian Hong\textsuperscript{1,2}},
  \textbf{Chen Cheng\textsuperscript{2}},
  \textbf{Quan Liu\textsuperscript{2}},
  \textbf{Yuhao Chen\textsuperscript{1}},
  \textbf{Enhong Chen\textsuperscript{1}}
\\
  \textsuperscript{1} University of Science and Technology of China,
  \textsuperscript{2} iFLYTEK Research Group,
\\
  \small{
\href{mailto:chencheng18@iflytek.com, jianhong@iflytek.com}{\faEnvelope[regular]\ \{jianhong, chencheng18\}@iflytek.com}
  }
}

\begin{document}
\maketitle
\begin{abstract}
Following complex instructions with multiple explicit constraints remains a fundamental challenge for large language models (LLMs). Existing alignment methods, such as DPO, optimize holistic reward signals that often under-emphasize strict satisfaction of individual constraints, particularly under out-of-distribution or multi-constraint settings. In this paper, we propose \methodname, a stage-wise optimization framework that decouples the alignment of subjective (soft) constraints from the optimization of objectively verifiable (hard) constraints. Stage 1 applies preference optimization with multiple negative samples to sharpen the sensitivity to soft constraints, while Stage 2 applies Reinforcement Learning with Verifiable Rewards (RLVR) to enforce strict compliance with hard constraints. To support this method, we construct \datasetname, a high-quality bilingual (English, Chinese) dataset of approximately 31{,}000 complex multi-constraint instructions. 
Extensive analyses validate the design of \methodname\ and show state-of-the-art performance on representative benchmarks against strong baselines, as well as genuine generalization.
We have released our code and dataset\footnote{\url{https://github.com/hj1993/STAIF}}.
\end{abstract}

\section{Introduction}
\label{sec:introduction}

Large language models (LLMs) have evolved into general-purpose assistants capable of handling diverse real-world tasks through natural language instructions. Users often specify multiple constraints simultaneously, spanning requirements on content, format, style, and more. 
In these settings, partial compliance is often insufficient; missing just one critical constraint can make the response unusable.

Despite progress in instruction tuning and preference alignment, reliable multi-constraint instruction following remains a fundamental challenge. Recent notable works such as Reinforcement Learning from Human Feedback (RLHF), Direct Preference Optimization (DPO), and Tulu3 \citep{christiano2017deep,ouyang2022training,rafailov2023dpo,lambert2024tulu3} typically optimize responses using a holistic preference or verification signal. While this paradigm is effective for overall helpfulness, it can obscure which individual constraints are satisfied or violated. A deeper issue is that constraints are heterogeneous in nature: subjective (soft) constraints (e.g., ``in an encouraging tone'') require comparative preference judgments, whereas objective (hard) constraints (e.g., ``output valid JSON'') demand deterministic, rule-based verification. Treating both types with a single holistic objective dilutes the binary correctness signal for hard constraints and fails to capture the nuanced preference distinctions needed for soft ones.

\begin{figure}[t]
  %\vspace{-0.6em}
  \centering
  %\includesvg[width=\linewidth]{figures/simple.svg}
  \includegraphics[width=\linewidth]{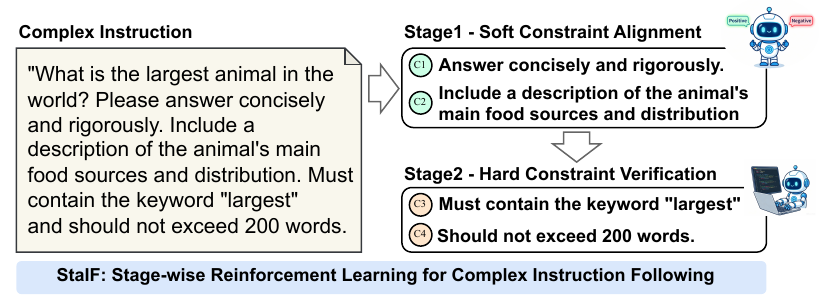}
  %\vspace{-0.8em}
  \caption{Illustration of \methodname: Stage 1 soft-constraint alignment; Stage 2 hard-constraint verification.}
  \label{fig:simple}
  \vspace{-1.5em}
\end{figure} 
Compounding these methodological challenges is the scarcity of high-quality, large-scale training data specifically designed for complex instruction following. Several benchmark datasets, such as IFEval, IFBench, and CFBench, have recently been introduced to evaluate instruction following abilities \citep{zhou2023ifeval,jiang2024,zhang2025cfbench}, but constructing corresponding training data at scale remains difficult due to inherent challenges in ensuring constraint compatibility, avoiding logical conflicts, etc. This shortage is particularly acute for Chinese language, where natively constructed complex-instruction datasets remain extremely limited.

To address these challenges, we propose Stage-wise Instruction Following (\textbf{\methodname}), a two-stage alignment framework that explicitly decouples soft-constraint preference alignment from hard-constraint verifiable optimization (Figure~\ref{fig:simple}). In Stage~1, we perform preference optimization to teach the model to distinguish responses that satisfy the intended soft constraints from those that omit or violate them, with a richer learning signal through multiple negative responses for better discrimination. In Stage~2, we apply Reinforcement Learning with Verifiable Rewards (RLVR) through Group Relative Policy Optimization (GRPO, \citet{shao2024deepseekmath}) to directly optimize for strict satisfaction of objectively hard constraints. This stage-wise design provides separate learning signals to different constraint types, enabling the model to improve both preference alignment and strict constraint compliance without trade-offs.

To support our stage-wise training approach, we develop an automated complex-constraint data synthesis framework that explicitly models task-constraint compatibility and filters invalid, redundant, or conflicting constraints during expansion. Using this framework, we construct \textbf{\datasetname}, a large-scale bilingual dataset (30\% English and 70\% Chinese) containing approximately 31,000 complex multi-constraint instructions. For each instruction, we generate high-quality positive responses, multiple negative responses according to the soft constraints, and rule-based verification code for hard constraints, making the dataset uniquely suitable for both preference learning and verifiable-reward reinforcement learning.

We evaluate \methodname\ on three instruction-following benchmarks: IFEval, IFBench, and CFBench. Experiments on two competitive open-source base models, Qwen2.5-7B-Instruct and Tulu3-8B-SFT, show consistent and significant performance improvements across all benchmarks. Notably, \methodname\ trained on Tulu3-8B-SFT achieves performance competitive with large scale models such as Gemini 2.5 on IFEval and IFBench despite being substantially smaller in model size. Ablation studies confirm that both training stages contribute meaningfully to the final performance. Experiments demonstrate that the model design of decoupling soft/hard constraints and using multiple negative sampling provides stronger contrastive learning signals. Fine-grained analysis further reflects genuine generalization rather than in-distribution memorization.

The key contributions of this paper are summarized as follows:
\begin{itemize}
  \setlength{\itemsep}{0pt}
  \setlength{\parsep}{0pt}
    \item We propose \methodname, a novel two-stage optimization framework with multiple negative samples that addresses the performance bottleneck of complex instruction following by decoupling soft-constraint preference alignment from hard-constraint verification.
    \item We propose an automated complex-constraint data synthesis framework and \datasetname, a large-scale high-quality bilingual dataset for complex instruction following.
    \item Extensive experiments on various instruction following benchmarks demonstrate state-of-the-art (SoTA) performance and strong generalization by \methodname.
\end{itemize}

\section{Related Work}
\label{sec:related-work}

\subsection{Complex Instruction Data Construction} 
Early instruction tuning works, including Alpaca  \footnote{\url{https://crfm.stanford.edu/2023/03/13/alpaca.html}}, Vicuna\footnote{\url{https://www.lmsys.org/blog/2023-03-30-vicuna/}}, and CAMEL \citep{li2023camel}, demonstrated the power of synthetic instruction data for open-source LLMs, but their datasets focused mainly on general instruction following or relatively simple constraints. To address this, Evol-Instruct pioneered depth/width instruction expansion using GPT-4 to generate more complex instructions \citep{xu2023wizardlm}, while Conifer proposed progressive learning for multi-level constraint following \citep{sun2024conifer}. AutoIF further automated the generation of instructions and verification code \citep{dong2025autoif}.
However, existing datasets suffer from: 1) a lack of explicit task-constraint compatibility modeling, leading to invalid or conflicting constraints; 2) insufficiently diverse negative samples for contrastive learning; and 3) scarcity of high-quality Chinese data. We construct \datasetname\ to address these issues.

\subsection{Preference-based Alignment Methods}
RLHF and DPO have become standard for LLM alignment, but their holistic reward signals can obscure individual constraint satisfaction, diluting learning signals in multi-constraint settings. Recent works such as MuSC \citep{huang2025musc} and IOPO \citep{iopo} enhance constraint sensitivity through multi-granularity self-contrastive training and input-output preference optimization, but still treat soft and hard constraints uniformly, failing to provide the deterministic signals required for strict hard-constraint compliance.
Step-by-Step Mastery focuses specifically on soft constraints, combining automatically constructed positive and negative samples with DPO and a curriculum organized by constraint quantity \citep{ren-etal-2025-step}. While these methods improve sensitivity to subjective requirements, they do not jointly enforce objectively verifiable hard constraints through a dedicated optimization stage.
In this paper, we apply preference alignment only to soft constraints and optimize hard constraints separately with a different approach. 

\subsection{Verification-based Methods}
%RLVR has proven to be effective for programmatically verifiable tasks such as mathematical reasoning and instruction following \citep{shao2024deepseekmath,lambert2024tulu3}. In instruction following, Tulu~3 applied RLVR with modest gains, while AutoIF and VerIF improved verification pipelines \citep{lambert2024tulu3,dong2025autoif,verif}. However, these works focus primarily on hard constraints and under-emphasize subjective soft preferences that are critical for user satisfaction. We design a two-stage framework to enhance for both soft and hard constraints.

RLVR has proven to be effective for programmatically verifiable tasks such as mathematical reasoning and instruction following \citep{shao2024deepseekmath,lambert2024tulu3}. In instruction following, Tulu~3 applied RLVR with modest gains, while AutoIF and VerIF improved verification pipelines \citep{lambert2024tulu3,dong2025autoif,verif}. More recently, reinforcement learning from checklist feedback derives instruction-specific criteria and combines AI judge and programmatic verifier scores into fine-grained rewards \citep{viswanathan2025checklists}. Beyond the Trade-off uses self-supervised reinforcement learning signals from reasoning models themselves to improve instruction following while preserving reasoning performance \citep{ren2025beyond}. Unlike these approaches, \methodname\ explicitly separates soft-constraint preference alignment from hard-constraint verifiable optimization.

\section{Preliminaries}
\label{sec:preliminaries}

In this section, we formally define the instruction-following setting and summarize common alignment objectives used throughout the paper.
Let $x$ denote an instruction and $y$ a model response. A policy $\pi_{\theta}(y \mid x)$ generates responses with parameters $\theta$, and $\pi_{\text{ref}}$ represents a base reference policy.

\subsection{DPO}
\label{sec:preliminaries:dpo}
DPO bypasses explicit reward modeling and optimizes $\pi_{\theta}$ directly from preferences as discussed in \citet{rafailov2023dpo}.
Under implicit reward parameterization, it yields the following maximum-likelihood objective:
\begin{equation}
\label{eq:dpo}
\begin{split}
&\mathcal{L}_{\text{DPO}}(\pi_\theta; \pi_{\text{ref}}) = \\
&- \mathbb{E}_{(x,y_w,y_l)\sim\mathcal{D}_p} \left[\log \sigma \left( \beta \log \tfrac{\pi_\theta(y_w \mid x)}{\pi_{\text{ref}}(y_w \mid x)} \right. \right. \\
&\left. \left. - \beta \log \tfrac{\pi_\theta(y_l \mid x)}{\pi_{\text{ref}}(y_l \mid x)} \right) \right]
\end{split}
\end{equation}
with an implicit reward form:
%The implicit reward can be written as
\begin{equation}
    r(x, y) = \beta \log \tfrac{\pi_{\theta}(y \mid x)}{\pi_{\text{ref}}(y \mid x)} + \beta \log Z(x),
    \label{eq:reward}
\end{equation}
where $Z(x)$ is a partition function.

\subsection{RLVR}
In settings with objectively checkable outcomes, reinforcement learning can use a deterministic verifier $V(x,y)\in\{0,1\}$ (or a scalar score) as reward, avoiding a learned reward model.
This verification reinforcement learning setup is commonly used when the correctness can be programmatically validated \citep{shao2024deepseekmath,verif}.
%(e.g., format constraints or exact-match tasks)
\subsection{GRPO}
GRPO estimates the advantages from a group of sampled completions $\{o_1,\dots,o_G\}$ for the same prompt and uses normalized group-relative rewards \citep{shao2024deepseekmath}.
Given rewards $\{r_i\}_{i=1}^{G}$, the normalized advantage is
\begin{equation}
    A_i = \frac{r_i - \mathrm{mean}(r_1,\dots,r_G)}{\mathrm{std}(r_1,\dots,r_G)}.
\end{equation}
\section{Data: \datasetname}
\label{sec:data}

\begin{figure*}[t]
  \centering
  \includegraphics[width=\linewidth]{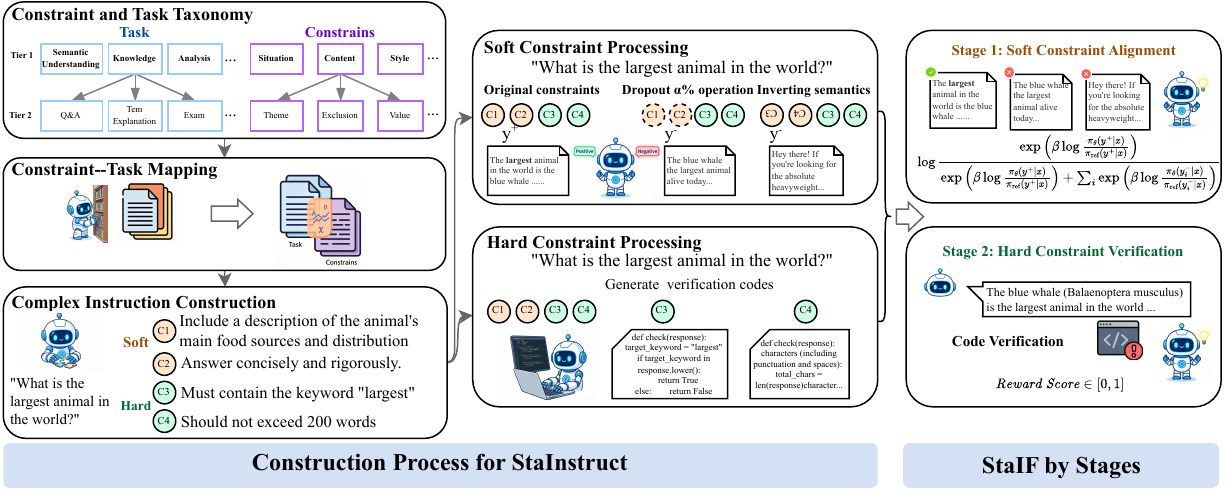}
  \caption{Overview of the data construction pipeline for \datasetname.}
  \label{fig:full}
\end{figure*}

In this section, we propose an automated complex-constraint data synthesis framework and construct \datasetname, a large-scale dataset designed for complex, multi-constraint instruction following tasks.
The dataset is constructed in both English and Chinese, addressing the lack of Chinese benchmarks that capture linguistic nuances and culturally specific constraints. 
Figure~\ref{fig:full} provides an overview of the data construction pipeline.

\subsection{Data Construction}
\label{sec:data:constr}
\paragraph{Taxonomy.} 
A complex instruction can be segmented into: a \emph{task} (e.g., ``write an essay'') and a set of \emph{constraints} (e.g., ``in Markdown format'').

To systematically ensure task diversity, constraint validity, and constraint complexity, we construct a taxonomy of both tasks and constraints, including the mapping between them. We group tasks into seven main categories, such as \textit{Semantic Understanding} and \textit{Analysis}, with 28 subcategories in total. For constraints, we adopt the comprehensive constraint type system of \citet{jiang2024} and \citet{iopo}, which contains five constraint categories (\textit{Content}, \textit{Situation}, \textit{Style}, \textit{Format}, \textit{Example}) and 26 subcategories.

\begin{figure}
  \centering
  \includegraphics[width=\linewidth]{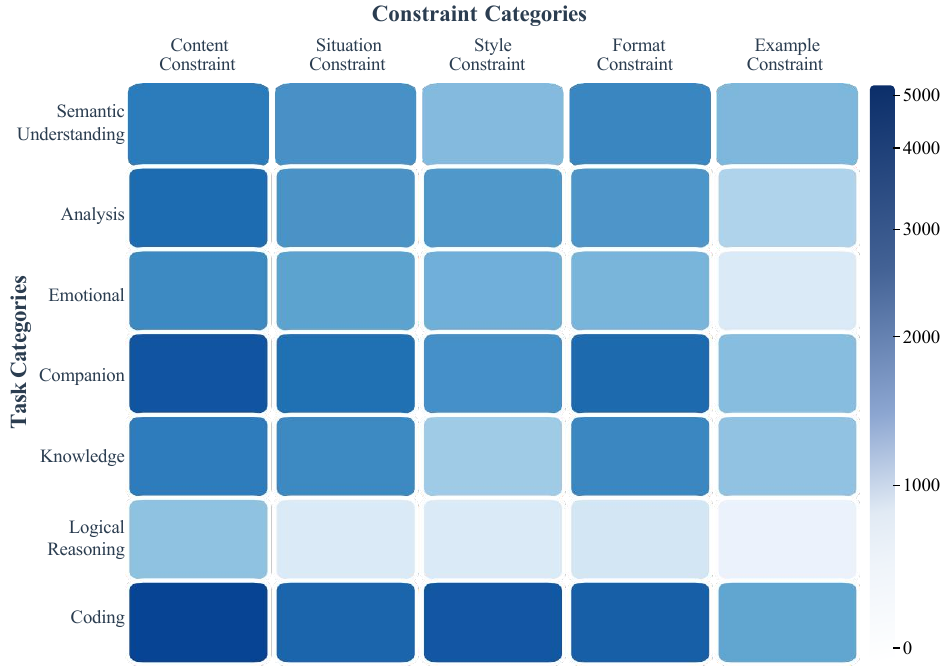}
  \caption{Mapping between task categories and constraint categories based on real-world user data.}
  \label{fig:heatmap}
  %\vspace{-1em}
\end{figure}

 We analyze real-world user data to map out the relationship between task categories and constraint categories (Figure~\ref{fig:heatmap}). It can be observed that for a task category, certain constraints are more likely to be posed by users than others. To ensure the validity of generated instructions in later steps, we restrict the allowable constraint categories for each task category accordingly. 

Detailed descriptions of the task and constraint taxonomies can be found in Appendix~\ref{sec:appendix-taxonomy}.

\paragraph{Instruction Expansion.} 
Based on the categorization, instructions, especially simple instructions, from our data pool are expanded through LLM inference to form an input set of complex instructions. A human-LLM collaborative quality check is in place to remove redundant, conflicting, or inappropriate constraints.

\paragraph{Constraint Processing.} 
%Structured decomposition and annotation between soft and hard constraint are conducted for each instruction. 
We first parse each instruction into a structured list of constraints, separating soft and hard constraints.

Following the algorithms in Section~\ref{sec:method}, for each constructed instruction $x$, we generate multiple negative samples $\{x_1^-, x_2^-,..., x_n^-\}$ by 1) removing one or more soft constraints mimicking the situation of missing constraints; or 2) inverting the semantics of soft constraints mimicking the situation of constraint violations.

For hard constraints, we leverage the LLM code generation capability to generate rule-based verification code to enable automated deterministic constraint satisfaction checks. Invalid or syntactically erroneous code is corrected by human-LLM collaborative revision. We perform a modest, feasible check by manually checking a random sample of generated verifier functions (200 checkers). If the accuracy falls below 95\%, we inform LLM and rerun the verifier generation process up to 4 times. Out of the generated final verifiers for around 1000 hard constraints, the accuracy is 96.3\%.
%To improve the alignment to soft constraints as described in \ref{sec:method}, for each constructed instruction $x$ with constraints $\{c_1, c_2,\dots\}$, multiple responses are generated by LLM prompting, including one possitive response $y^+$ and multiple negative responses $\{y^-_1,y^-_2,\dots,y^-_n\}$.
%The preferred response $y^+$ fully follows the exact set of constraints outlined in $x$, while each negative response $y^-_i$ is generated by either removing or reverse one or more soft constraints.
\paragraph{Response Generation and Verification.}
For each instruction $x$, we prompt an LLM to generate a set of responses, including a positive response $y^+$ that fully follows the exact set of constraints outlined in $x$, and multiple negative responses $\{y_1^-,y_2^-,...,y_n^-\}$ generated from the negative sample set $\{x_1^-, x_2^-,..., x_n^-\}$. 

To validate the selected $y^+$, we follow previous work on automatic instruction-following verification and preference construction \citep{verif,iopo} by combining rule-based checks for objective constraints with model-based judgments paired with human review for subjective constraints.
%Specifically, we first parse each instruction into a structured list of constraints and apply an \emph{objective verifier} to deterministically check rule-based constraints (e.g., word count, Markdown/JSON validity, required fields).
%We then apply a \emph{semantic verifier} (LLM-as-a-judge with rubric prompts) to assess subjective constraints such as tone, relevance, and situational compliance.
%Among candidate responses, we select the response that passes all objective checks and achieves the highest semantic-verifier score as $y^{+}$; the remaining candidates form dispreferred responses.
If no candidate satisfies all objective constraints, we regenerate responses or discard the instruction during filtering.

We use specialized LLMs for the aforementioned constructing steps to ensure data construction quality. For instruction expansion, constraint processing, and response generation/verification, we use GPT-4.1\footnote{\url{https://openai.com/index/gpt-4-1/}}
 for English and Doubao 1.8\footnote{\url{https://seed.bytedance.com/blog/official-release-of-seed1-8-a-generalized-agentic-model}} for Chinese. For code generation, we use Doubao 1.8.

\subsection{Dataset Statistics.}
\begin{figure}[!hptb]
  \centering
  \includegraphics[width=0.95\linewidth]{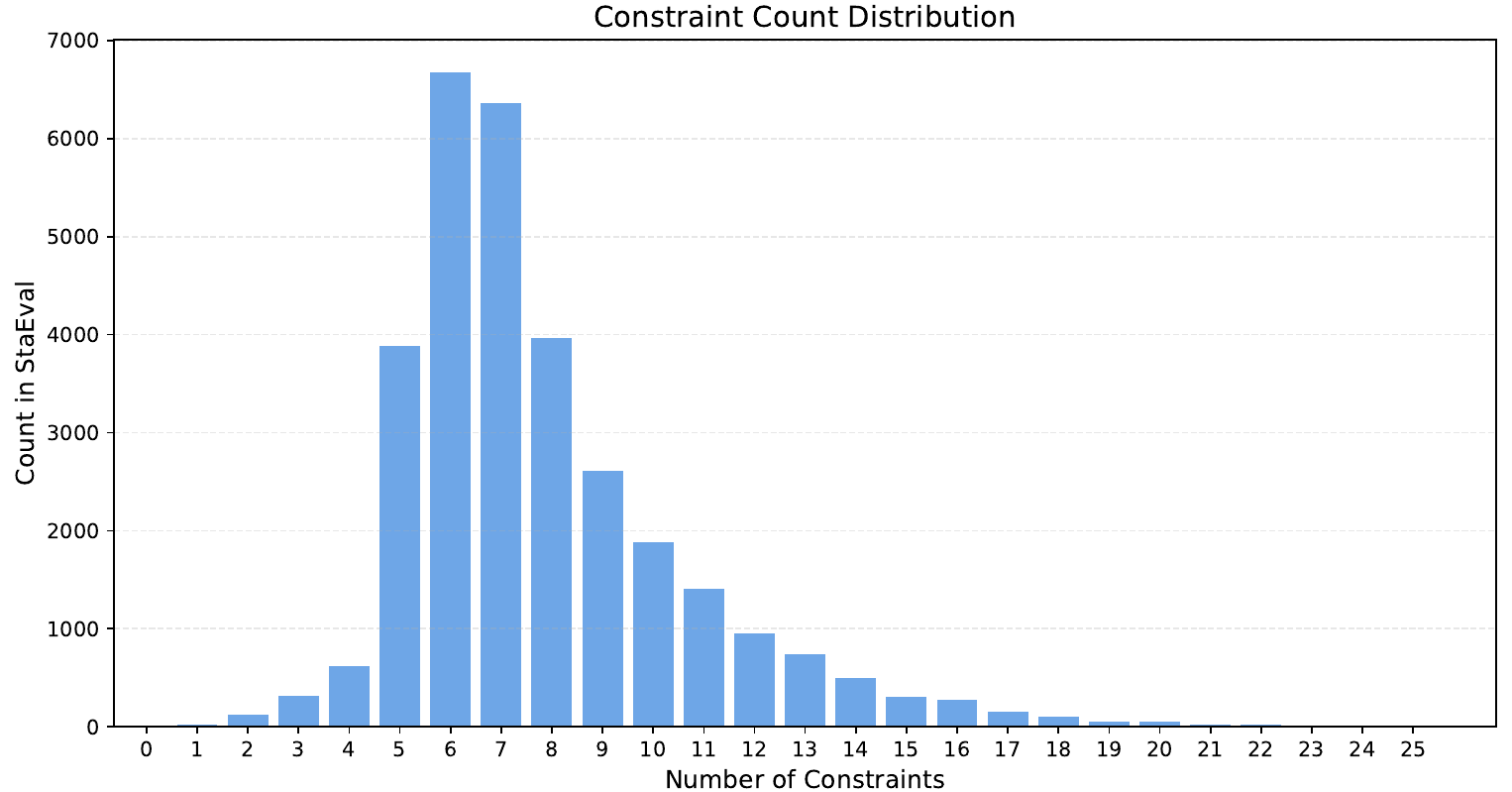}
  \caption{Distribution of the number of constraints per instruction in \datasetname.}
  \label{fig:constraint-dist}
  %\vspace{-1em}
\end{figure}

\begin{figure}[t]
  \centering
  \includegraphics[width=0.95\linewidth]{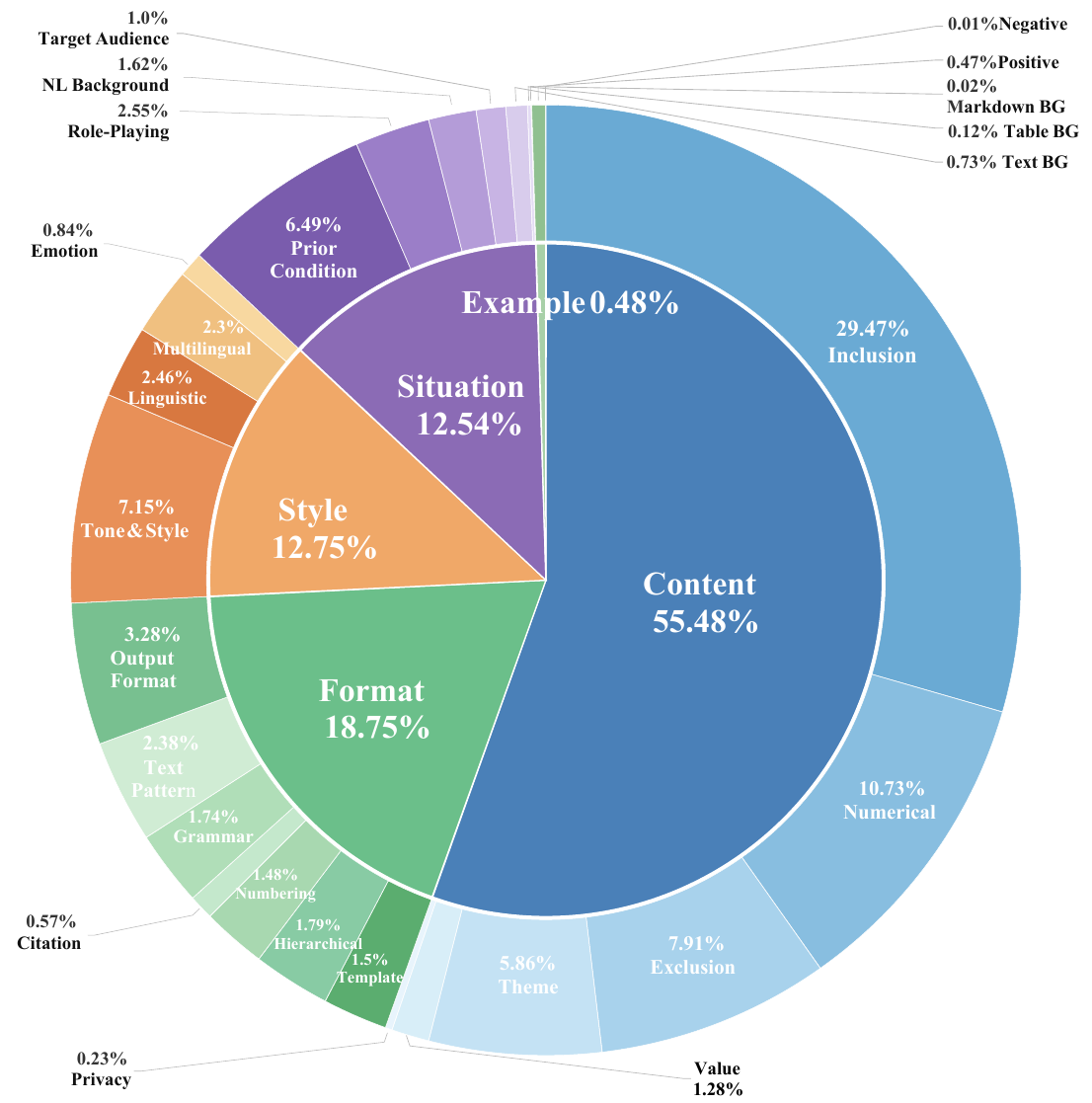}
  \caption{Constraint type distribution in \datasetname.}
  \label{fig:constraint-pie}
\end{figure}

The constructed dataset, \datasetname, contains 31{,}088 instructions in total, among which 70\% are in Chinese and 30\% in English.
The number of constraints per instruction ranges from 1 to 29, with an average of 7.8.
Figure~\ref{fig:constraint-dist} shows the distribution of constraint count. Around 47\% of the constraints are soft constraints, while 53\% are hard constraints.  
Figure~\ref{fig:constraint-pie} summarizes the constraint-type composition of \datasetname, with the inner circle representing the distribution of the five major categories as described in Section~\ref{sec:data:constr} and the outer circle representing that of the sub-categories.

\section{Method}
\label{sec:method}
In this section, we propose \methodname, a stage-wise optimization algorithm designed for complex instruction following.

We assume that an instruction $x$ consists of a set of constraints $C=\{c_1,c_2,\dots\}$. 
Our complete optimization pipeline has two sequential stages: 1) preference alignment for soft constraints, and 2) RLVR for hard constraints \citep{shao2024deepseekmath,verif}, decoupling the optimization for the two types of constraints with customized paradigms and training objectives.

\paragraph{Stage 1: Preference Alignment for Soft Constraints.}
The core optimization objective of this stage is to enable the model to achieve precise alignment with the implicit, subjective preference-based constraints embedded in complex instructions.

In view of the core characteristics of subjective constraints, which are the absence of absolute quantitative criteria and the reliance on human preference judgment, we adopt a DPO-style objective~\citep{rafailov2023dpo} rather than conventional RLHF. Standard RLHF requires a separate reward model and PPO optimization, which can be costly and unstable. DPO directly optimizes the policy from pairwise preferences through an implicit reward function, providing a simpler and more stable way to capture fine-grained preference signals.

As described in Section~\ref{sec:data}, for each instruction $x$ in \datasetname, we construct one preferred response $y^+$ and multiple dispreferred responses $\{y^-_1,y^-_2,\dots,y^-_n\}$.
Our setting corresponds to a ``winner-takes-all'' choice among multiple candidates, where only the top choice (the preferred response that satisfies the exact set of constraints of the instruction as discussed in Section~\ref{sec:data}) is preferred and the rest (dispreferred responses) are unranked ``losers''.
We therefore adopt the Plackett--Luce model \citep{plackett1975,luce2012}, and derive our preference expression as follows:
\begin{equation}
\label{eq:p}
\begin{split}
    &p(y^+ \succ \{y^-_1,\dots,y^-_n\}\mid x)\\
    =&\frac{\exp(r(x,y^+))}{\exp(r(x,y^+))+\sum_{i=1}^n \exp(r(x,y^-_i))},
\end{split}
\end{equation}
where $r(x,y)$ is the reward function in Equation~\ref{eq:reward}.

We then construct the corresponding maximum likelihood objective function using the aforementioned preference function for multi-candidate preference probabilities:
%and concurrently introduce the native implicit Kullback-Leibler (KL) divergence regularization term from the DPO framework:
\begin{equation}
\label{eq:staif}
\begin{split}
    &\mathcal{L}_{\mathrm{StaIF}}(\pi_{\theta}; \pi_{\text{ref}})\\
    = & -\mathbb{E}_{(x, \mathbf{y}) \sim \mathcal{D}} \left[\log S_\mathbf{y}\left(\beta \log\frac{\pi_{\theta}(y^+ \mid x)}{\pi_{\text{ref}}(y^+ \mid x)}\right)\right],
\end{split}
\end{equation}
 where $\mathbf{y}=\{y^+, y^-_1,\dots,y^-_n\}$, $S(x)$ is the \emph{softmax} function, and $\beta$ controls the strength of the implicit KL regularization. 

 In particular, our objective function is mathematically equivalent to InfoNCE loss~\citep{oord2018cpc}, a foundational objective in contrastive learning. This connection provides a theoretical justification for our use of multiple negative samples: increasing the number of negatives tightens the mutual information lower bound optimized by InfoNCE, leading to more robust learning of constraint satisfaction patterns. This is consistent with the experimental results in Section~\ref{sec:experiments:further}.
 
Detailed derivations of the above can be found in Appendix~\ref{app:dev}.

\paragraph{Stage 2: Verification for Hard Constraints.}
In this stage, we optimize objective constraint satisfaction using RLVR, addressing a core limitation of existing preference optimization methods: insufficient learning signals for hard constraints. The rewards in RLVR are computed by deterministic checkers \citep{zhou2023ifeval,verif} through the following steps: 1) For each hard constraint $c_j^H$ of instruction $x$, construct the verification function $f_j(x,y)$ based on the rule verification code generated in Section~\ref{sec:data:constr}. $f_j(x,y) = 1$ if $y$ meets constraint $c_j^H$, and $f_j(x,y) = 0$ otherwise. 2) The reward function is constructed as the average of the verification functions of all hard constraints in $x$: $R(x,y)=\mathrm{avg}_j(f_j(x,y))$.

With the reward function, we then adopt the GRPO algorithm as the core reinforcement learning optimization framework for this stage \citep{shao2024deepseekmath}. Compared with the conventional PPO algorithm, GRPO eliminates the need for an additional value network, and is optimally suited for the large-model fine-tuning scenario based on deterministic rule-based rewards in this study.

\section{Experiments}
\label{sec:experiments}

\subsection{Experimental Setup}
Three representative instruction-following benchmarks are used to evaluate our method's performance: \textbf{IFEval}, \textbf{IFBench}, and \textbf{CFBench}. For IFEval and IFBench, we report prompt-level and instruction-level accuracy  under both strict and loose matching.
For CFBench, we report constraint satisfaction rate (CSR), instruction satisfaction rate (ISR), and priority satisfaction rate (PSR) to provide a comprehensive view of constraint-following performance.

We train \methodname\ on two base models: \textbf{Qwen2.5-7B-Instruct} \citep{hui2024qwen2} and \textbf{Tulu3-8B-SFT} (with RL, \citet{lambert2024tulu3}).
These models represent competitive instruction-tuned baselines from different model families, allowing us to assess the generalizability of our approach. Detailed implementation settings are given in Appendix~\ref{appendix:exp}.

We compared \methodname\ against a broad set of strong open-source and proprietary baselines across the benchmarks, including frontier large/mid-scale models: GPT-4.1, DeepSeek-v3.2 \citep{deepseek2025v32}, Gemini 2.5 \citep{comanici2025gemini}, Doubao 1.8, and QwQ-32B\footnote{\url{https://qwenlm.github.io/blog/qwq-32b/}} as general reference, and models with comparable model size for comparison: Llama3.1-8B-Instruct \citep{grattafiori2024llama}, Tulu3-8B \citep{lambert2024tulu3}, Crab-7B-DPO \citep{qi2025constraint}, and VerIF \citep{verif}.
For the two base models used to train \methodname, we conduct controlled comparisons by evaluating the same base models trained, respectively, under SFT \citep{wei2021finetuned} and DPO \citep{rafailov2023dpo} with our training data \datasetname.

\subsection{Main Results}
\label{sec:experiments:main}

\begin{table*}[t]
  \centering
  \setlength{\tabcolsep}{3.5pt}
  \small
  \begin{tabular}{lccccccccccc}
    \toprule
    \multirow{2}{*}{\textbf{Model}}
      & \multicolumn{4}{c}{\textbf{IFEval}}
      & \multicolumn{4}{c}{\textbf{IFBench}}
      & \multicolumn{3}{c}{\textbf{CFBench}} \\
      \cmidrule(lr){2-5} \cmidrule(lr){6-9} \cmidrule(lr){10-12}
    & \textbf{P-S} & \textbf{P-L} & \textbf{I-S} & \textbf{I-L}
    & \textbf{P-S} & \textbf{P-L} & \textbf{I-S} & \textbf{I-L}
    & \textbf{CSR} & \textbf{ISR} & \textbf{PSR} \\
    \midrule
    %\multicolumn{12}{l}{\textbf{Base models}} \\
    GPT-4.1 & 87.8 & 91.2 & 91.1 & 93.2 & 38.7 & 43.8 & 42.1 & 47.5 & 83.0 & 61.0 & 70.0 \\
    DeepSeek-v3.2 & 86.4 & 90.5 & 90.8 & 93.5 & 39.7 & 43.5 & 42.3 & 46.8 & 87.0 & 65.0 & 72.0 \\
    Gemini 2.5 & 89.8 & 91.1 & 92.9 & 94.0 & 39.7 & 50.3 & 43.0 & 52.5 & 83.0 & 60.0 & 70.0\\
    Doubao 1.8 & 83.4 & 87.0 & 88.5 & 90.0 & 29.3 & 36.4 & 32.8 & 39.7 & 83.0 & 60.0 & 68.0\\
    QwQ-32B & 83.5 & 87.7 & 88.8 & 91.8 & 36.3 & 40.4 & 39.7 & 44.2 & 87.0 & 61.0 & 70.0 \\
    \midrule
    LLaMa3.1-8B-Instruct & 64.8 & 69.3 & 74.6 & 78.1 & 22.8 & 26.9 & 25.1 & 31.0 & 58.0 & 20.0 & 26.0 \\
    Tulu3-8B & 78.2 & 80.6 & 84.6 & 86.9 & 23.1 & 27.2 & 25.7 & 29.9 & 74.0 & 35.0 & 47.0 \\
    Crab-7B-DPO & 47.7 & 57.7 & 59.9 & 68.1 & 15.9 & 21.8 & 17.6 & 24.5 & 54.0 & 17.0 & 23.0\\
    VerIF & 84.5 & 87.1 & 89.3 & 91.4 & 23.5 & 24.8 & 25.9 & 28.1 & 72.0 & 33.0 & 43.0\\
    \midrule
    %\multicolumn{12}{l}{\textbf{StagewiseRL (before/after)}} \\
    Qwen2.5-7B-Instruct & 69.4 & 72.6 & 78.3 & 81.0 & 24.4 & 27.5 & 27.7 & 30.7 & 80.0 & 44.0 & 56.0 \\
    \quad \text{+SFT} & 75.8 & 79.1 & 82.0 & 84.8 & 29.2 & 34.7 & 32.5 & 38.5 & 81.0 & 44.0 & 54.0 \\
    \quad \text{+DPO} & 78.1 & 81.6 & 83.8 & 86.8 & 35.2 & 38.7 & 36.3 & 40.6 & 80.0 & 45.0 & 56.0 \\
    \rowcolor{gray!15}\quad \textbf{+ \methodname} & \textbf{87.0} & \textbf{88.7} & \textbf{91.0} & \textbf{92.3} & \textbf{44.2} & \textbf{48.3} & \textbf{46.0} & \textbf{49.9} & \textbf{82.0} & \textbf{51.0} & \textbf{61.0} \\
    Tulu3-8B-SFT & 68.3 & 70.7 & 75.7 & 78.0 & 22.1 & 23.5 & 22.7 & 25.1 & 69.0 & 29.0 & 40.0 \\
    \quad \text{+SFT (secondary)} & 77.4 & 81.3 & 83.8 & 87.0 & 28.6 & 34.0 & 31.9 & 37.0 & 73.0 & 33.0 & 46.0\\
    \quad \text{+DPO} & 78.2 & 81.6 & 84.1 & 86.5 & 32.9 & 37.1 & 34.1 & 38.1 & 70.0 & 32.0 & 41.0 \\
    \rowcolor{gray!15}\quad \textbf{+ \methodname} & \textbf{90.4} & \textbf{91.1} & \textbf{92.9} & \textbf{93.5} & \textbf{50.3} & \textbf{52.0} & \textbf{53.1} & \textbf{55.2} & \textbf{73.0} & \textbf{35.0} & \textbf{45.0} \\
    \bottomrule
  \end{tabular}
  \caption{Main results on IFEval, IFBench, and CFBench. For IFEval and IFBench we report accuracy (\%) at the \textbf{prompt} (P) and \textbf{instruction} (I) levels under \textbf{strict} (S) and \textbf{loose} (L) matching: \textbf{P-S}, \textbf{P-L}, \textbf{I-S}, \textbf{I-L}. For CFBench, we report \textbf{CSR}, \textbf{ISR}, and \textbf{PSR} (\%).} 
  \label{tab:full-results}  
\end{table*}

Experimental results are in Table~\ref{tab:full-results}. \methodname\ demonstrates strong performance across all benchmarks. 

\paragraph{Performance on Par with Large-scale Models.} Notably, \methodname\ trained on Tulu3-8B-SFT achieves competitive or even superior performance compared to models with significantly larger parameter sizes on IFEval and IFBench. Specifically, %on both benchmarks, Gemini 2.5 gives the best performance among the mid/large-scale baselines. 
on IFEval, \methodname\ on Tulu3-8B-SFT achieves very similar performance with Gemini 2.5, both achieving an average score of $92.0\%$. On the more challenging IFBench, \methodname\ achieves an average score of $52.7\%$, outperforming Gemini 2.5 ($46.4\%$) by a notable margin of $+6.3\%$. These results demonstrate the effectiveness of our approach in improving instruction-following capabilities.
%and indicate that \methodname\ substantially strengthens robustness on harder instruction-following cases rather than only improving easier constraints.

\paragraph{SoTA Performance among Similar-size Models.} \methodname\ establishes a clear and consistent advantage among models of comparable parameter scale. On IFEval and IFBench, \methodname\ on Tulu3-8B-SFT achieves SoTA performance, while on CFBench, \methodname\ on Qwen2.5-7B-Instruct achieves SoTA performance. Specifically, \methodname\ on Tulu3-8B-SFT outperforms Tulu3-8B by an average margin of $+9.4\%$ on IFEval, $+26.2\%$ on IFBench, while \methodname\ on Qwen2.5-7B-Instruct outperforms all models of similar size by an average margin of at least $+12.7\%$.

\paragraph{Advantage over Single Training Paradigms.} 
In the last part of Table~\ref{tab:full-results}, where the two base models are trained with the same data under different paradigms, we observe that DPO generally yields better performance than SFT. Comparing \methodname\ to DPO, we observe consistent performance improvements of $+7.2\%/+11.5\%/+17.0\%$ across three benchmarks on Qwen2.5-7B-Instruct, and of $+9.4\%/+17.1\%/+3.3\%$ on Tulu3-8B-SFT. 

We also compare \methodname\ with VerIF which optimizes both types of constraints jointly in a single stage \citep{verif}. \methodname\ achieves average performance improvements of $+3.9\%$ on IFEval, $+27.1\%$ on IFBench, and $+1.7\%$ on CFBench, justifying the advantage of treating soft and hard constraints in separate stages. For comparability, we use the results under Tulu3-8B-SFT for both models. We select VerIF as a most-recent representative of alignment methods. Further comparison of \methodname\ with other recent related works as discussed in Section~\ref{sec:related-work} can be found in Appendix~\ref{appendix:add}. Results demonstrate consistent gains over self-supervised, checklist-based, curriculum-based, and self-contrastive alignment approaches.

These results support our choice of design of decoupling the alignment of soft constraints from optimization on hard constraints.

\subsection{Further Analysis}
\label{sec:experiments:further}
\paragraph{Ablation Study.}
To understand the contribution of each training stage, we conduct ablation experiments by removing Stage 1 (w/o stage1) and Stage 2 (w/o stage2) in \methodname\ respectively.

\begin{table*}[t]
  \centering
  \setlength{\tabcolsep}{3.5pt}
  \small
  \begin{tabular}{llccccccccccc}
\toprule
\multirow{2}{*}{\textbf{Model}} & \multirow{2}{*}{\textbf{Method}} 
  & \multicolumn{4}{c}{\textbf{IFEval}} 
  & \multicolumn{4}{c}{\textbf{IFBench}} 
  & \multicolumn{3}{c}{\textbf{CFBench}} \\
\cmidrule(lr){3-6} \cmidrule(lr){7-10} \cmidrule(lr){11-13}
& & \textbf{P-S} & \textbf{P-L} & \textbf{I-S} & \textbf{I-L} 
  & \textbf{P-S} & \textbf{P-L} & \textbf{I-S} & \textbf{I-L} 
  & \textbf{CSR} & \textbf{ISR} & \textbf{PSR} \\
\midrule
\multirow{3}{*}{Qwen2.5-7B}
& \textbf{\methodname}  & \textbf{87.0} & \textbf{88.7} & \textbf{91.0} & \textbf{92.3} & \textbf{44.2} & \textbf{48.3} & \textbf{46.0} & \textbf{49.9} & \textbf{82.0} & \textbf{51.0} & \textbf{61.0} \\
  & w/o stage1         & 80.2 & 82.8 & 86.3 & 88.1 & 37.8 & 41.2 & 39.7 & 43.9 & 80.0 & 45.0 & 58.0 \\
  & w/o stage2         & 79.4 & 82.0 & 85.5 & 87.6 & 38.4 & 42.5 & 40.3 & 45.4 & 81.0 & 46.0 & 58.0 \\
\midrule
\multirow{3}{*}{Tulu3-8B}
  & \textbf{\methodname}    & \textbf{90.4} & \textbf{91.1} & \textbf{92.9} & \textbf{93.5} & \textbf{50.3} & \textbf{52.0} & \textbf{53.1} & \textbf{55.2} & \textbf{73.0} & \textbf{35.0} & \textbf{45.0} \\
  & w/o stage1         & 85.1 & 87.0 & 89.3 & 90.2 & 42.2 & 45.6 & 43.3 & 47.8 & 70.0 & 33.0 & 41.0 \\
  & w/o stage2         & 80.1 & 83.0 & 85.8 & 88.2 & 35.0 & 38.7 & 37.3 & 41.5 & 72.0 & 33.0 & 44.0 \\
\bottomrule
\end{tabular}
\caption{Ablation study on the two training stages of \methodname, evaluated on Qwen2.5-7B and Tulu3-8B.}
\label{tab:ablation}
\end{table*}
\begin{table*}[t]
  \centering
  \setlength{\tabcolsep}{7pt}
  \small
  \begin{tabular}{lcccccccc}
    \toprule
    \multirow{2}{*}{\textbf{Method}}
      & \multicolumn{4}{c}{\textbf{IFEval}}
      & \multicolumn{4}{c}{\textbf{IFBench}} \\
    \cmidrule(lr){2-5} \cmidrule(lr){6-9}
    & \textbf{P-S} & \textbf{P-L} & \textbf{I-S} & \textbf{I-L}
    & \textbf{P-S} & \textbf{P-L} & \textbf{I-S} & \textbf{I-L} \\
    \midrule
    DPO (joint) & 79.1 & 82.0 & 84.6 & 86.9 & 36.9 & 39.5 & 39.7 & 43.7 \\
    RLVR (joint) & 81.6 & 83.3 & 86.9 & 89.4 & 41.3 & 43.2 & 44.1 & 45.1 \\
    DPO+RLVR (joint) & 84.5 & 86.8 & 88.4 & 90.7 & 41.1 & 45.1 & 44.7 & 48.3 \\
    \rowcolor{gray!15}\textbf{\methodname} & \textbf{87.0} & \textbf{88.7} & \textbf{91.0} & \textbf{92.3} & \textbf{44.2} & \textbf{48.3} & \textbf{46.0} & \textbf{49.9} \\
    \bottomrule
  \end{tabular}
  \caption{Impact of decoupling soft and hard constraints, evaluated on Qwen2.5-7B.}
  \label{tab:joint-optimization}
\end{table*}

Results are reported in Table~\ref{tab:ablation}. Both stages contribute meaningfully to the overall performance, and removing either one consistently degrades the performance across all benchmarks and both base models. For Qwen2.5-7B, Stage 1 brings an average performance gain of $+5.4\%$, $+6.5\%$, and $+3.7\%$ on IFEval, IFBench, and CFBench, respectively, while Stage 2 separately contributes $+6.1\%$, $+5.5\%$, and $+3.0\%$. On Tulu3-8B, the average performance gain is $+4.1\%$, $+7.9\%$, and $+3.0\%$ for Stage 1, and $+7.7\%$, $+14.5\%$, and $+1.3\%$ for Stage 2. The largest improvements are observed on the out-of-distribution IFBench benchmark, where \methodname\ achieves up to a $+15.8\%$ performance gain by adding Stage 2, demonstrating that decoupling soft and hard constraint optimization significantly improves generalization to unseen instructions.

\paragraph{Impact of Decoupling Soft and Hard Constraints.} \methodname\ treats alignment of soft constraints and optimization on hard constraints differently in separate stages. We compared \methodname\ performance with methods applying DPO jointly on all constraints, RLVR jointly on all constraints, and DPO+RLVR jointly on all constraints, as shown in Table~\ref{tab:joint-optimization}. It can be observed that \methodname\ achieves the best performance across all metrics. While the two-stage method applying DPO+RLVR jointly on all constraints yields a better performance comparing to simply stage, \methodname\ manages to improve the performance even further by 1.3\% to 3.2\%. These results indicate that decoupling the alignment of soft constraints from
optimization on hard constraints is more effective than treating them jointly within a unified objective.

\paragraph{Effect of Multiple Negative Samples.} \methodname\ employs multiple negative samples during training to provide a richer contrastive learning signal. In Section~\ref{sec:method}, we provide theoretical support for this design choice. To validate it empirically, we compare \methodname, as well as the model with only soft-constraint alignment (Stage 1 only), against the variant trained with single negative sample.

\begin{table*}[t]
  \centering
  \setlength{\tabcolsep}{3.5pt}
  \small
  \begin{tabular}{lllccccccccccc}
\toprule
\multirow{2}{*}{\textbf{Model}} & \multirow{2}{*}{\textbf{Method}} & \multirow{2}{*}{\textbf{Negative }}
  & \multicolumn{4}{c}{\textbf{IFEval}}
  & \multicolumn{4}{c}{\textbf{IFBench}}
  & \multicolumn{3}{c}{\textbf{CFBench}} \\
\cmidrule(lr){4-7} \cmidrule(lr){8-11} \cmidrule(lr){12-14}
& &\textbf{Sample} & \textbf{P-S} & \textbf{P-L} & \textbf{I-S} & \textbf{I-L}
  & \textbf{P-S} & \textbf{P-L} & \textbf{I-S} & \textbf{I-L}
  & \textbf{CSR} & \textbf{ISR} & \textbf{PSR} \\
\midrule
\multirow{4}{*}{Qwen2.5-7B}
  & \multirow{2}{*}{Stage 1 only} & Single  & 76.3 & 80.4 & 83.5 & 86.5 & 34.0 & 38.1 & 36.1 & 40.6 & 80.0 & 43.0 & 56.0 \\
  &  & Multiple & 79.4 & 82.0 & 85.5 & 87.6 & 38.4 & 42.5 & 40.3 & 45.4 & 81.0 & 46.0 & 58.0 \\
  & \multirow{2}{*}{\textbf{\methodname}} & Single & 85.0 & 86.3 & 89.7 & 90.6 & 42.2 & 46.5 & 44.9 & 47.5 & 80.0 & 46.0 & 56.0 \\
 &  & \textbf{Multiple} & \textbf{87.0} & \textbf{88.7} & \textbf{91.0} & \textbf{92.3} & \textbf{44.2} & \textbf{48.3} & \textbf{46.0} & \textbf{49.9} & \textbf{82.0} & \textbf{51.0} & \textbf{61.0} \\
\midrule
\multirow{4}{*}{Tulu3-8B}
  & \multirow{2}{*}{Stage 1 only} & Single & 76.5 & 79.4 & 83.5 & 86.1 & 31.6 & 34.6 & 33.7 & 37.0 & 69.0 & 30.0 & 40.0 \\
  &  & Multiple & 80.1 & 83.0 & 85.8 & 88.2 & 35.0 & 38.7 & 37.3 & 41.5 & 72.0 & 33.0 & 44.0 \\
  & \multirow{2}{*}{\textbf{\methodname}} & Single & 87.5 & 88.3 & 91.0 & 91.7 & 48.3 & 50.7 & 50.2 & 52.8 & 72.0 & 33.0 & 43.0 \\
& & \textbf{Multiple} & \textbf{90.4} & \textbf{91.1} & \textbf{92.9} & \textbf{93.5} & \textbf{50.3} & \textbf{52.0} & \textbf{53.1} & \textbf{55.2} & \textbf{73.0} & \textbf{35.0} & \textbf{45.0} \\
\bottomrule
\end{tabular}
\caption{Comparison of single vs. multiple negative samples in \methodname, evaluated on Qwen2.5-7B and Tulu3-8B.}
\label{tab:negative_samples}
\end{table*}
\begin{table*}[t]
  \centering
  \setlength{\tabcolsep}{3.5pt}
  \small
  \begin{tabular}{lcccc}
    \toprule
    \textbf{Model} & \textbf{MT-Bench} & \textbf{GSM8K} & \textbf{MMLU-Pro} & \textbf{BBH} \\
    \midrule
    Qwen2.5-7B-Instruct  & 7.4 & 91.6 & 55.9 & 72.3 \\
    \rowcolor{gray!15}\textbf{+\methodname} & 7.5 & 91.4 & 56.0 & 74.4 \\
    \midrule
    Tulu3-8B-SFT & 6.3 & 78.7 & 35.0 & 67.8 \\
    \rowcolor{gray!15}\textbf{+\methodname} & 7.1 & 81.3 & 35.4 & 71.2 \\
    \bottomrule
  \end{tabular}
  \caption{Semantic quality evaluation before and after applying \methodname. MT-Bench is reported on its standard 10-point scale; GSM8K, MMLU-Pro, and BBH are reported as accuracy percentages.}
  \label{tab:semantic-quality}
\end{table*}

Table~\ref{tab:negative_samples} shows that using multiple negative samples consistently outperforms the single negative variants across both base models and all benchmarks. This performance improvement holds for both the full \methodname\ model and Stage 1 only, showing that the benefit is not stage-sensitive. This analysis validates our design choice: exposing the model to diverse negative examples during training provides a richer learning signal, enabling it to better discriminate between constraint-satisfying and constraint-violating responses. 

\paragraph{Granular Analysis on Constraint Types.} To better understand how our method generalizes across different distributions of instruction constraints, we break down performance by constraint type on IFEval and IFBench. We compare the Tulu3-8B base model, Tulu3-8B+Stage 1, and Tulu3-8B+\methodname.

\begin{figure}[t]
  \centering
  \includegraphics[width=\linewidth]{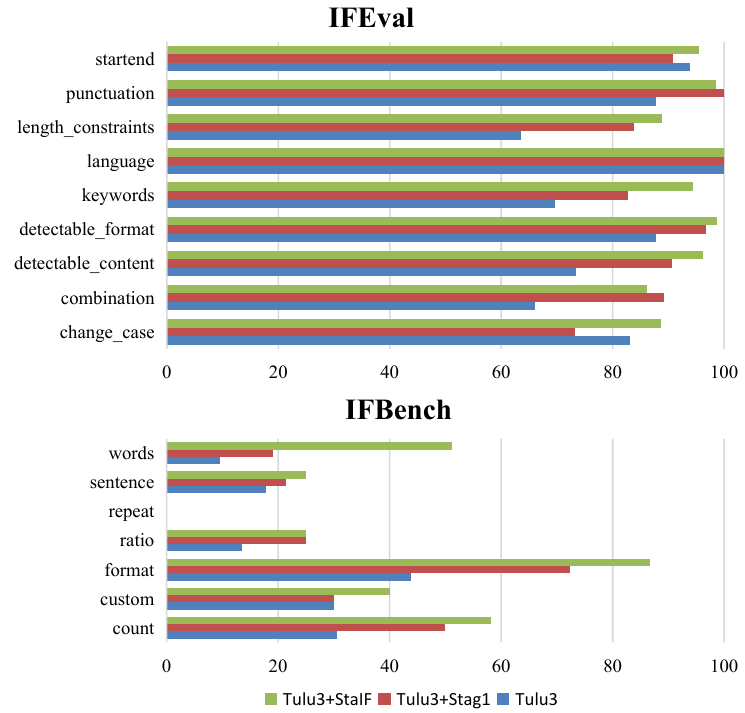}
  \caption{Performance by constraint type on IFEval and IFBench (prompt-level, loose accuracy) for Tulu3, Tulu3+Stage 1, and Tulu3+\methodname.}
  \label{fig:type}
  %\vspace{-0.8em}
\end{figure}

Figure~\ref{fig:type} shows that \methodname\ yields broad and consistent gains across most IFEval constraint categories. %This trend is expected, as the constraints in our training data are closely aligned with those in IFEval. 
Similar improvements are also observed on IFBench. This is particularly noteworthy given that constraints in IFBench are drawn from a substantially different distribution than those in the training data. The observed improvements on most constraint types in IFEval and IFBench serve as evidence of genuine generalization rather than in-distribution memorization.

\paragraph{Semantic Quality.} We test \methodname\ performance on several general task benchmarks: MT-Bench \citep{mtbench} for conversation ability, GSM8K\footnote{\url{https://github.com/openai/grade-school-math}} for math, MMLU-Pro \citep{MMLU} for knowledge, and BBH \citep{BBH} for inference. Table~\ref{tab:semantic-quality} shows that \methodname\ preserves or improves general response quality and reasoning performance, indicating that \methodname\ does not sacrifice semantic quality to satisfy rule-based checkers in Stage 2.  

\section{Conclusion}
\label{sec:conclusion}

We propose \methodname, a stage-wise framework that separates soft-constraint alignment from hard-constraint optimization, combining preference learning with multiple negative samples and RL with verifiable rewards. We also construct \datasetname, a large-scale bilingual dataset of complex multi-constraint instructions, under an automated complex-constraint data synthesis framework.

Our methodology achieve SoTA performance on representative benchmarks with different languages and constraint distributions, demonstrating the efficacy of \methodname\ and the potential of RL in complex instruction following. In future work, we plan to explore further methodology and data enhancement such as stricter product-based reward instead of partial-credit reward, as well as broader application scenarios.

\section*{Limitations}
\label{sec:limitations}

In the training dataset, we set the number of negative samples to be three. The three negative samples are constructed respectively through 1) dropping 50\% of the soft constraints; 2) dropping all the soft constraints; and 3) inverting the semantics of all the soft constraints. While the experiment results show effectiveness of using multiple negative samples, the number of negative samples and the construction scheme are not data driven or tuned to achieve better performance. 

Testing results in Section~\ref{sec:experiments} are based on single-run. While the consistent direction of improvement across all metrics and large margin on several benchmarks justify that the performance improvement is not a noise-driven effect, multiple runs would further strengthen the evaluation.

\section*{Ethical Considerations}
\label{sec:ethical}

This work investigates complex instruction-following capabilities of LLMs. For data collection, we utilize real-world user queries that have been thoroughly cleaned to remove all personally identifiable information (PII) as well as toxic or sensitive content. We also employ LLMs for both data generation and automatic evaluation of experimental results. Given that, we acknowledge that our approach inherits the known risks associated with LLMs, including potential risk of bias and fairness \citep{anthis-etal-2025-impossibility,gallegos-etal-2024-bias}. We mitigate these risks by manually checking the quality of a random sample of the results. 

We use AI assistance (Claude and ChatGPT) to refine some wordings.

\bibliography{custom}
\appendix
\clearpage
\onecolumn
\section{Taxonomies}
Detailed categorization and description for task and constraint:
\label{sec:appendix-taxonomy}
\begin{table}[!htbp]
\centering
\small
\begin{tabular}{@{}p{0.13\textwidth} p{0.19\textwidth} p{0.61\textwidth}@{}}
\toprule
\textbf{Category} & \textbf{Sub-category} & \textbf{Description}\\
\midrule
\multirow[t]{6}{=}{Semantic Understanding} & Semantic Analysis & Analyze the logical structure, semantic relationships, and core meaning of text.\\
 & Sentiment Classification & Identify the emotional tendency or intensity of text.\\
 & Element Extraction & Extract key information units (e.g., people, events, relationships) from unstructured text.\\
 & Element Recognition & Identify extracted entities (e.g., ``Apple'' $\rightarrow$ ``Organization'', ``Beijing'' $\rightarrow$ ``Location'').\\
 & Natural Language Inference & Determine the logical relationship among contents (Entailment/Contradiction/Neutral).\\
 & Text Processing & Preprocessing (cleaning, segmentation, normalization) or postprocessing (formatting, filtering) of text data.\\
\midrule
\multirow[t]{3}{=}{Analysis} & Content Analysis & Generate analysis of certain topic or area (e.g., industry, products, culture) as instructed.\\
 & Content Review & Provide subjective evaluations, critical comments, or feedback as instructed.\\
 & Information Inquiry & Provide consulting (e.g., skills) to user queries.\\
\midrule
\multirow[t]{3}{=}{Emotional Companion} & Suggestion & Offer practical solutions or actionable advice for user life problems/scenarios.\\
 & Recommendation & Recommend items, content, or services tailored to user preferences or needs.\\
 & Casual Conversation & Engage in informal, natural dialogue to fulfill social or emotional needs.\\
\midrule
\multirow[t]{3}{=}{Knowledge} & Knowledge Q\&A & Answer factual, conceptual, or procedural questions.\\
 & Term Explanation & Define professional terms, clarify connotations, and illustrate usage scenarios.\\
 & Exam & Solve academic/examination questions (multiple-choice, subjective, computational) across disciplines.\\
\midrule
\multirow[t]{4}{=}{Logical Reasoning} & Symbolic Reasoning & Perform logical deductions using mathematical symbols, formal languages, or logical operators.\\
 & Common Sense Reasoning & Infer conclusions based on universal daily life knowledge and intuitive judgment.\\
 & Relational Reasoning & Deduce the connection between entities (e.g., causal, hierarchical, associative relationships).\\
 & Conditional Reasoning & Derive potential consequences or conclusions based on given premises or hypothetical conditions.\\
\midrule
\multirow[t]{5}{=}{Coding} & Code Generation & Convert natural language descriptions into executable code snippets or full programs.\\
 & Code Completion & Predict and fill in incomplete code (e.g., function names, syntax, logical blocks) during coding.\\
 & Code Debugging \& Fixing & Identify syntax errors, logical bugs, or performance issues and provide corrected code.\\
 & Code Refactoring & Optimize code structure (readability, maintainability, efficiency) without changing core functionality.\\
 & Code Explanation & Clarify the purpose, logical flow, and implementation details of existing code.\\
\midrule
\multirow[t]{4}{=}{Writing} & Content Creation & Generate original content (articles, stories, reports, etc.) based on user requirements or themes.\\
 & Summary Generation & Condense long text into a concise summary that retains key information and logical relationships.\\
 & Translation & Convert text between different languages while preserving semantic accuracy, style, and cultural nuances.\\
 & Content Rewriting & Revise existing content to improve clarity, fluency, tone, or adapt to specific usage scenarios.\\
 \bottomrule
\end{tabular}
\caption{Task Taxonomy.}
\label{tab:task_taxonomy}
\end{table}
\begin{table}[H]
\centering
\small
\begin{tabular}{@{}p{0.12\textwidth} p{0.18\textwidth} p{0.62\textwidth}@{}}
\toprule
\textbf{Category} & \textbf{Sub-category} & \textbf{Description}\\
\midrule
\multirow[t]{9}{=}{Content} & Theme & The generated content should focus on a specific topic or field.\\
 & Exclusion & Clearly specify the information or content that should not be included in the generated content.\\
 & Inclusion & Clearly specify the particular information or content that must be included in the generated content.\\
 & Value & The generated content should not contain information that violates values, such as safety, false information, discrimination, or bias.\\
 & Numerical & Limit the length and number of words, sentences, and paragraphs in the generated content, or use numerical precision constraints to ensure accuracy.\\
 & Target Audience & The generated content should target a specific audience, which affects terminology, detail level, and complexity.\\
 & Prior Condition & When a specific intention is met, a particular process should be followed to perform an operation or output specific content.\\
 & Natural Language Process Background & Add natural-language process information, such as procedures or business processes, to assist in generating answers.\\
 & Markdown Process Background & Add markdown-formatted process information, such as procedures or business processes, to assist in generating answers.\\
\midrule
\multirow[t]{2}{=}{Situation} & Table Background Information & Background information is presented in table form, providing a series of markdown-formatted tables to assist in generating answers.\\
 & Text Background Information & Background information is presented in text form, providing textual background information to assist in generating answers.\\
\midrule
\multirow[t]{4}{=}{Style} & Tone and Style & The generated content should adopt a specific tone and style, such as formal, polite, academic, concise, literary, romantic, or sci-fi.\\
 & Emotion & The generated content should express a specific emotion or mood, such as being positive, inspiring, or empathetic.\\
 & Linguistic Characteristics & Use specific linguistic features, such as metaphors, personification, and other rhetorical devices.\\
 & Multilingual & The content should be generated in a specific language or switch between languages according to complex patterns.\\
 \midrule
\multirow[t]{7}{=}{Format} & Output Format & The generated content should be in a specific data format, such as tables, JSON, HTML, LaTeX, or Markdown.\\
 & Text Pattern & Use specified fonts, font sizes, or special emojis to satisfy formatting requirements.\\
 & Grammar Structure & The generated content should strictly follow specific grammatical structures, such as subject--predicate--object or subject--verb patterns.\\
 & Citation & The generated content should include citations to sources and follow specific citation or reference styles.\\
 & Numbering and List & The generated content should use numbered lists or bullet points to organize information.\\
 & Hierarchical Structure & The generated content should be organized according to a specific hierarchical structure, such as headings and subheadings.\\
 & Template & The generated content should follow a specific layout or format, such as text alignment, paragraph indentation, and introduction--body--conclusion templates.\\
\midrule
\multirow[t]{2}{=}{Example} & Positive Example & Provide examples that meet the requirements and require the model to generate content based on these examples.\\
 & Negative Example & Provide examples that do not meet the requirements and require the model to avoid generating similar content.\\
 \bottomrule
\end{tabular}
\caption{Constraint Taxonomy.}
\label{tab:constraint_taxonomy}
\end{table}
\twocolumn

\clearpage
\section{Mathematical Derivations}
\label{app:dev}
\subsection{Derivation of Preference Function in Equation~\ref{eq:p}}
The Plackett-Luce model is a probabilistic framework for modeling rankings over a set of discrete alternatives, derived from Luce's choice axiom \citep{luce2012} and Plackett's permutation analysis \citep{plackett1975}.
The general form of Plackett-Luce model is 
\begin{equation}
p(\tau | \mathbf{y}) = \prod_{k=1}^K \frac{f(y_{\tau(k)})}{\sum_{j=i}^K f(y_{\tau(j)})}
\label{eq:pl_general_full_ranking}
\end{equation}
where $\mathbf{y}=\{y_1,\dots,y_K\}$, and $\tau$ represents a permutation of $\mathbf{y}$. The general form expresses the probability of a complete ranking. However, in our work, it is a partial ranking case where we only consider the top-ranked option $y_{\tau(1)}$ while the rest are unranked alternatives. Thus the Plackett-Luce model simplifies to:
\begin{equation}
p(y_{\tau(1)} \succ \mathbf{y}\backslash y_{\tau(1)}| \mathbf{y}) =  \frac{f(y_{\tau(k)})}{\sum_{j=i}^K f(y_{\tau(j)})}
\label{eq:pl_general_partial_ranking}
\end{equation}
Let positive response $y^+$ be $y_{\tau(1)}$, negative responses $\{y_1^-,\dots,y_n^-\}$ be $\mathbf{y}\backslash y_{\tau(1)}$, and replace the function $f$ in the equation above with DPO score function $\exp(r(x,y)))$, we get our preference function as shown in Equation~\ref{eq:p}:
\begin{equation}
\begin{split}
    &p(y^+ \succ \{y^-_1,\dots,y^-_n\}\mid x)\\
    =&\frac{\exp(r(x,y^+))}{\exp(r(x,y^+))+\sum_{i=1}^n \exp(r(x,y^-_i))},
\end{split}
\end{equation}

\begin{figure*}[!b]
\begin{equation}
\label{eq:loss}
\begin{aligned}
    &\mathcal{L}_{\text{StaIF}}(\pi_{\theta}; \pi_{\text{ref}})\\
    =& - \mathbb{E}_{(x, \mathbf{y}) \sim \mathcal{D}} \left[ \log \frac{\exp\left( \beta \log\frac{\pi_{\theta}(y^+ \mid x)}{\pi_{\text{ref}}(y^+ \mid x)}\right)}{\exp\left( \beta \log\frac{\pi_{\theta}(y^+ \mid x)}{\pi_{\text{ref}}(y^+ \mid x)}\right)+\sum_{i} \exp\left( \beta \log\frac{\pi_{\theta}(y^-_i \mid x)}{\pi_{\text{ref}}(y^-_i \mid x)}\right)}\right]\\
    =& - \mathbb{E}_{(x, \mathbf{y}) \sim \mathcal{D}} \left[\log S_\mathbf{y}\left(\beta \log\frac{\pi_{\theta}(y^+ \mid x)}{\pi_{\text{ref}}(y^+ \mid x)}\right)\right]
\end{aligned}
\end{equation}
%\vspace{0.75\baselineskip}
\begin{equation}
\label{eq:nce}
\mathcal{L}_{\text{InfoNCE}} = -\mathbb{E}_{(x, y^+, \{y_i^-\}_{i=1}^n) \sim \mathcal{D}} \left[ \log \frac{\exp\left( \frac{s(x, y^+)}{t} \right)}{\exp\left( \frac{s(x, y^+)}{t} \right) + \sum_{i=1}^n \exp\left( \frac{s(x, y_i^-)}{t} \right)} \right]
\end{equation}
\end{figure*}
\subsection{Derivation of Maximum Likelihood Loss in Equation~\ref{eq:staif}}
With the preference function derived above, and plugging in the softmax function $S_\mathbf{y}(y_i)=\tfrac{\exp(y_i)}{\sum_j\exp(y_i)}$, we get the maximum likelihood loss function as shown in Equation~\ref{eq:loss}.

\subsection{InfoNCE Equivalence}
The canonical InfoNCE loss is defined in Equation~\ref{eq:nce}, where $x$ is the \textit{anchor sample}, $y^+$ is the \textit{positive sample} that is semantically paired with the anchor, $\{y_i^-\}_{i=1}^n$ is a set of \textit{negative samples} that are not paired with the anchor, $s(x, y): \mathcal{X} \times \mathcal{Y} \to \mathbb{R}$ is a similarity function measuring the compatibility between anchor $x$ and sample $y$, and $t > 0$ is the temperature parameter that controls the sharpness of the probability distribution over candidates.

In our study, the ``similarity'' between an instruction $x$ and a response $y$ is represented by the reward function $r(x, y) = \log \frac{\pi_\theta(y|x)}{\pi_{\text{ref}}(y|x)}$. Substituting the reward model into the canonical InfoNCE loss in Equation~\ref{eq:nce}, and replacing $t$ with $\beta=1/t$ as the reverse temperature, which is a common thermodynamic interpretation of temperature in probabilistic models, we get our loss function shown in Equation~\ref{eq:staif} and Equation~\ref{eq:loss}.
This equivalence provides a principled explanation for our experimental result that multiple negative samples outperform single negative samples (Table 3). From the contrastive learning perspective:
\begin{itemize}
    \item The InfoNCE loss optimizes a lower bound on the mutual information $I(x; y^+)$ between instructions and their correct constraint-satisfying responses
    \item Increasing the number of negative samples $n$ tightens this mutual information lower bound, leading to more robust learning of constraint satisfaction patterns
    \item This effect is particularly pronounced in out-of-distribution settings (e.g., IFBench), where stronger generalization requires tighter mutual information estimation
\end{itemize}

\clearpage
\section{Experimental Details}
\label{appendix:exp}
\subsection{Training Setup.} 
We apply the same training recipe to both base models, Qwen2.5-7B-Instruct and Tulu3-8B-SFT. All training experiments are implemented using the open-source ms-swift framework\footnote{\url{https://pypi.org/project/ms-swift/}} on NVIDIA A800 GPUs. Our approach employs a two-stage training pipeline, with hyperparameters detailed as follows.

\begin{table}[h]
\centering
\small
\begin{tabular}{ll}
\toprule
\multicolumn{2}{c}{\textbf{Stage 1}} \\
\midrule
Initial learning rate & $1 \times 10^{-5}$ \\
Number of training epochs & 3 \\
$\beta$ & 0.1 \\
Maximum sequence length & 4{,}096 \\
Per-GPU batch size & 1 \\
\bottomrule
\end{tabular}
\end{table}

\begin{table}[h]
\centering
\small
\begin{tabular}{ll}
\toprule
\multicolumn{2}{c}{\textbf{Stage 2}} \\
\midrule
Initial learning rate & $1 \times 10^{-6}$ \\
Maximum sequence length & 4{,}096 \\
Number of training epochs & 2 \\
Number of rollouts per iteration & 16 \\
KL divergence loss coefficient & $1 \times 10^{-3}$ \\
\bottomrule
\end{tabular}
\end{table}
We use IFEval as the validation set to select the best checkpoint \citep{lambert2024tulu3} and train the models on \datasetname\ with early stopping if there is no improvement of performance on IFEval
for more than 3 checkpoints.
\subsection{Evaluation Setup.}
\paragraph{Baseline Models.}
We compare our proposed approach against a comprehensive set of large language models (LLMs), including both large- and small-scale, closed- and open-source models.
For large-scale models (GPT-4.1, Doubao 1.8, DeepSeek v3.2, and Gemini 2.5), we evaluate via official API endpoints. For small-scale models (Tulu3-8B, Llama-3.1-8B-Instruct, QwQ-32B, Crab-7B-DPO), we serve the models through vLLM 0.21.0 \footnote{\url{https://github.com/vllm-project/vllm}} with a temperature setting of $0.0$ for high-throughput inference to ensure deterministic and reproducible outputs. 

For IFEval and IFBench, deterministic rule-based checkers are used for scoring, and for CFBench evaluations are conducted using GPT-4.1 as the judge model. GPT-4.1 is a design choice since it has been widely adopted in recent LLM research for its high inter-annotator agreement and alignment with human preferences. Although GPT-4.1 is also used in the data generation process, the fact that it does not directly generate the responses, and that performance margin on CFBench using GPT-4.1 as the judge model is not larger than other non-judged benchmarks, justify low risk of judge/data-construction coupling.

\section{Experimental Details}
\label{appendix:add}

\paragraph{Comparison with Recent Alignment Methods.}
Table~\ref{tab:recent-methods} compares \methodname\ with four recent alignment methods discussed in Section~\ref{sec:related-work} on their available reported metrics: VeRL \citep{ren2025beyond}, RLCF \citep{viswanathan2025checklists}, Step \citep{ren-etal-2025-step}, and MuSC \citep{huang2025musc}. These results demonstrate consistent gains over self-supervised, checklist-based, curriculum-based, and self-contrastive alignment approaches.

\begin{table}[t]
  \centering
  \setlength{\tabcolsep}{3.5pt}
  \small
  \begin{tabular}{lccccccc}
    \toprule
    \multirow{2}{*}{\textbf{Model}} 
    & \multicolumn{4}{c}{\textbf{IFEval}} 
    & \multicolumn{3}{c}{\textbf{CFBench}} \\
    \cmidrule(lr){2-5} \cmidrule(lr){6-8}
    & \textbf{P-S} & \textbf{P-L} & \textbf{I-S} & \textbf{I-L} 
    & \textbf{CSR} & \textbf{ISR} & \textbf{PSR} \\
    \midrule
    \multicolumn{8}{l}{\textbf{Qwen2.5-7B-Instruct}}\\
    \midrule
    VeRL & -- & 83.9* & -- & -- & 76.0* & 40.0* & 52.0* \\
    RLCF  & 72.6* & 77.3* & 80.3* & 84.1* & -- & -- & -- \\
    \rowcolor{gray!15}\textbf{\methodname} & \textbf{87.0} & \textbf{88.7} & \textbf{91.0} & \textbf{92.3} & \textbf{82.0} & \textbf{51.0} & \textbf{61.0} \\
    \midrule
    \multicolumn{8}{l}{\textbf{LLaMA3-8B-Instruct}}\\
    \midrule
    Step  & 72.5* & 77.1* & 80.3* & 84.1* & -- & -- & -- \\
    MuSC & -- & -- & -- & -- & 69.0* & 30.0* & 42.0* \\
    \rowcolor{gray!15}\textbf{\methodname}  & \textbf{88.1} & \textbf{88.5} & \textbf{90.6} & \textbf{91.2} & \textbf{75.0} & \textbf{36.0} & \textbf{48.0} \\
    \bottomrule
  \end{tabular}
  \caption{Comparison with recent instruction-following alignment methods. Dashes indicate metrics not reported by the corresponding method.\\
  * are cited from the corresponding reference for each method.}
  \label{tab:recent-methods}
\end{table}
%\clearpage
\section{Prompts}
\label{appendix:prompt}

\begin{promptbox}{Constraint Extraction Prompt}
You are an expert in NLP and constraint checking. Your task is to analyze a given instruction and identify which constraints need to be checked. The instruction contains a specific task query along with several explicitly stated constraints. Based on the instruction, return a list of checker names that should be applied to these constraints.

## Example 1
**Instruction:** Write a 300+ word summary of the Wikipedia page "https://en.wikipedia.org/wiki/Raymond_III_Count_of_Tripoli". Do not use any commas and highlight at least 3 sections that have titles in markdown format, for example *highlighted section part 1*, *highlighted section part 2*, *highlighted section part 3*.

**Response:** NumberOfWordsChecker: 300+ words `<sep>` HighlightSectionChecker: highlight at least 3 sections that have titles in markdown format `<sep>` ForbiddenWordsChecker: Do not use any commas.

## Example 2
**Instruction:** Analyze the sentiment of the following review and perform sentiment attribution. Present the results in JSON format with fields "emotion" and "sentiment attribution".

**Response:** FormatChecker: results must be in JSON format `<sep>` JSONFieldChecker: JSON must contain fields "emotion" and "sentiment attribution".
---
## [Instruction]
### Your task:
- Generate checker names and constraint descriptions: Extract and generate appropriate checker names and their corresponding constraint descriptions from the instruction.
- Use `<sep>` to separate outputs: Checker names and descriptions must be separated by `<sep>`.
- Only focus on explicitly stated constraints: Only process constraints clearly written in the instruction.
- Ensure constraint descriptions are complete and scope is clearly defined: For example, if a 300-word requirement is involved, clearly specify whether the limit applies to the entire text or a specific part.
- Do not generate checkers for the task itself or its quality: Only check constraints, not the core task content or its completion quality.
- Keep language consistent: If the instruction is in Chinese/English, output the constraint descriptions in the corresponding language.
- Each checker is responsible for exactly one constraint: Strictly follow the principle of "one checker per constraint".
- Do not output constraints not present in the instruction: Only process constraints explicitly given in the instruction.
- If there are no constraints, output only `<null>`: If the instruction contains no constraints, output only `<null>`.
\end{promptbox}

\begin{promptbox}{Hard/Soft Constraint Classification Prompt}
You are tasked with implementing a Python function `check_following` that determines whether a given `response` satisfies the constraint defined by a checker. The function should return `True` if the constraint is satisfied, and `False` otherwise.

## Example
**##Instruction:**
Task: Generate a short summary of the text in at most 32 words. Additionally, I have a specific constraint for your output: in your entire response, please refrain from the use of any commas.

Text: The Swiss will turn 35 during the Rio Games, where he will play in the singles and mixed doubles, although he has not decided on the men's doubles. [...]

**##Checker:** NumberOfWordsChecker: summary must be in at most 32 words
**##response:**
```python
def check_following(response):
    words = response.split()
    return len(words) <= 32
```
---
## Requirements
- The function should be self-contained with necessary imports.
- Do NOT use `nltk`.
- Only return exactly Python code, without any other information.
- The code only judges the constraint in the Checker, not other constraints in the instruction.
- All code must be inside the `check_following` function.
- The function name must be `check_following`.
---
**##Instruction:** [instruct]
**##Checker:** [check]
**##response:**
\end{promptbox}

\begin{promptbox}{Soft Constraint Removal Prompt}
## Role
You are an efficient data processing expert specializing in simplifying instruction constraints.

## Task
Analyze the instruction below, which consists of a "core task" and multiple "constraints". Your task is to:
1. Identify and retain the core task.
2. Remove the specified constraints while keeping all other constraints unchanged.
3. Strictly maintain the same language as the original instruction: if the original is in Chinese, output in Chinese; if in English or another language, output in that language.
4. Preserve the original format (core task first, followed by remaining constraints in list form).

## Restrictions
- Output only the simplified instruction, without any explanation or additional commentary.
---
**#Instruction:** {text}
**#Constraints to remove:** {constraints}
**#Simplified instruction:**
\end{promptbox}

\begin{promptbox}{Soft Constraint Negation Prompt}
## Role
You are an efficient data processing expert specializing in constructing contrastive instructions through "constraint negation" while preserving the core task objective.

## Task
Analyze the instruction below, which contains a "core task" and "constraints". Based on the provided constraint, generate a negated instruction according to the following rules:
1. Preserve the core task: Strictly retain the original intent and goal of the core task without altering its essence.
2. Precisely negate the constraint: Identify and reverse the modifying constraints, for example:
   - Tone: "formal" -> "casual and colloquial"
   - Length: "brief" -> "detailed and lengthy"
   - Audience: "target users are adults" -> "target users are children"
3. Logical consistency: The negated requirement must be logically executable and free of contradictions.
4. Language consistency (highest priority): The output language must match the input instruction exactly.
   - If the input is in Chinese, the output must be in Chinese.
   - If the input is in English, the output must be in English.
   - Do not translate the instruction into another language.
5. Format consistency: The core task comes first, followed by the negated constraints in list form.

## Example
**Input:** Please summarize this article. Constraint: use a formal tone.
**Output:** Please summarize this article. Constraint: use a humorous and casual tone.
---
**#Instruction:** {text}
**#Constraint:** {constraints}
**#Negated instruction:**
\end{promptbox}

\begin{promptbox}{Code Generation Prompt}
You are tasked with implementing a Python function `check_following` that determines whether a given `response` satisfies the constraint defined by a checker. The function should return `True` if the constraint is satisfied, and `False` otherwise.

## Example
**##Instruction:**
Task: Generate a short summary of the text in at most 32 words. Additionally, I have a specific constraint for your output: in your entire response, please refrain from the use of any commas.

Text: The Swiss will turn 35 during the Rio Games, where he will play in the singles and mixed doubles, although he has not decided on the men's doubles. [...]

**##Checker:** summary must be in at most 32 words
**##response:**
```python
def check_following(response):
    words = response.split()
    return len(words) <= 32
```
---
## Requirements
- The function should be self-contained with necessary imports.
- Do NOT use `nltk`.
- Only return exactly Python code, without any other information.
- The code only judges the constraint in the Checker, not other constraints in the instruction.
- All code must be inside the `check_following` function.
- The function name must be `check_following`.

Please generate code based on the given instruction and checker.
---
**##Instruction:** [instruct]
**##Checker:** [check]
**##response:**
\end{promptbox}
\end{document}